\documentclass[pdflatex,sn-basic,iicol]{sn-jnl}

\usepackage{graphicx}%
\usepackage{multirow}%
\usepackage{amsmath,amssymb,amsfonts}%
\usepackage{amsthm}%
\usepackage{mathrsfs}%
\usepackage[title]{appendix}%
\usepackage{xcolor}%
\usepackage{textcomp}%
\usepackage{manyfoot}%
\usepackage{booktabs}%
\usepackage{algorithm}%
\usepackage{algorithmicx}%
\usepackage{algpseudocode}%
\usepackage{listings}%
\usepackage{natbib}
\begin{document}

\title[Article Title]{Concepts Learned Visually by Infants Can Contribute to Visual Learning and Understanding in AI Models}

\author*{\fnm{Shify} \sur{Treger}}\email{shify.treger@weizmann.ac.il}

\author{\fnm{Shimon} \sur{Ullman}}\email{shimon.ullman@weizmann.ac.il}

\affil{\orgdiv{Faculty of Mathematics and Computer Science}, \orgname{Weizmann Institute of Science}, \orgaddress{\city{Rehovot}, \country{Israel}}}

\abstract{Early in development, infants learn to extract surprisingly complex aspects of visual scenes. This early learning comes together with an initial understanding of the extracted concepts, such as their implications, causality, and using them to predict likely future events. In many cases, this learning is obtained with little or no supervision, and from relatively few examples, compared to current network models. Empirical studies of visual perception in early development have shown that in the domain of objects and human-object interactions, early-acquired concepts are often used in the process of learning additional, more complex concepts. In the current work, we model how early-acquired concepts are used in the learning of subsequent concepts, and compare the results with standard deep network modeling. We focused in particular on the use of the concepts of animacy and goal attribution in learning to predict future events in dynamic visual scenes. We show that the use of early concepts in the learning of new concepts leads to better learning (higher accuracy) and more efficient learning (requiring less data), and that the combination of early and new concepts shapes the representation of the concepts acquired by the model and improves its generalization. 
We further compare advanced vision-language models to a human study in a task that requires an understanding of the behavior of animate vs. inanimate agents, with results supporting the contribution of early concepts to visual understanding. We finally briefly discuss the possible benefits of incorporating aspects of human-like visual learning into computer vision models.}

\keywords{Infant Visual Learning; Infant Social Learning; Computational Cognitive Modeling; Computer Vision; Vision-Language Models; Artificial Intelligence}



\maketitle

\section{Introduction}\label{Intro}

In early development, infants learn a broad range of useful concepts and visual tasks, which can be challenging from a computational point of view. Our goal in this work is to model aspects of this intriguing capacity and to compare human vision to vision-language models (VLMs) (We use the term VLMs because our focus is on vision-related aspects; however, our discussion also applies to so-called large multimodal models (LMMs).) that do not have this capacity. Finally, we consider the incorporation of this early human vision capacity in future large models. 
 
 In the domains of objects and human-object interactions (on which we focus), infants learn during the first year of life to recognize hands, their configuration, and their interactions with objects \citep{woodward1998infants, gergely2002rational, saxe2005secret}, a task where significant and meaningful features can be non-salient and highly variable and therefore difficult to learn. 

Infants learn, in an unsupervised manner, to perform figure-ground segmentation, which took years and a great effort to develop \citep{kirillov2023segment}. In the domain of dealing with other people, starting at 3-6 months of age, infants learn to detect and follow the gaze of another person and establish joint attention, based on head orientation and later on eye direction \citep{scaife1975capacity, flom2017gaze, d1997demonstration}. This task is difficult because ‘gaze’ is not a physical entity that appears explicitly in the image, and  cues for gaze direction can be subtle and difﬁcult to extract and use.  

The learning of early visual concepts comes together with an initial understanding of aspects of the meaning of concepts, in terms of their implications, causality, or use to predict probable future events. For example, in learning to identify hands, infants also learn that hands cause other objects to move and change location \citep{saxe2005secret, kiraly2003early}. All of this learning is obtained in many cases with little or no supervision, and from relatively few examples, compared to current models \citep{ullman2019model}. In learning about objects and human-object interactions, early learned concepts are often used in the process of acquiring additional related concepts. For example, the relation of in front/behind appears to be a prerequisite to subsequent learning of containment \citep{ullman2019model}. Learning to identify the direction of gaze is used to predict future actions \citep{falck2006infants}, and it later plays a role in the development of communication and language \citep{tomasello2009cultural}. 

In the current work, we model how early-acquired visual concepts are used in the learning of subsequent concepts. In particular, we focus on using the concepts of animacy and goal attribution in learning to predict future events in dynamic scenes. The main approach is to compare the learning of new concepts in two similar networks, where only one of the two uses a decomposition, similar to what was shown in infant studies, of learning a visual task by learning a simple concept first and then using it for learning the new task. For example, we compare the representation of the concept ‘animate’ and how it is used by the model, in the two different learning schemes: learning ‘animate’ first and then using it to predict future behavior, compared to combined learning of animacy together with prediction of future behavior. 

We refer to the version that uses human-like decomposition of concepts as the ‘cognitive’ model, and the standard, end-to-end training, as the ‘naive’ model. We compare the cognitive and standard approaches along two main directions: one is the performance of the trained models, and the other is a comparison of what was learned and represented in the two types of model. Briefly, in terms of model performance, the results show that the use of early concepts in the learning of new concepts leads to better learning (higher accuracy) and more efficient learning (requiring less data). In terms of the learned concept representations, the results show that when the concepts are learned in a human-like manner, the emerging representation is more useful, as measured in terms of generalization to novel data and tasks. We next examine the performance of a range of current large models in tasks that involve the use of animacy and goal attribution in predicting future human behavior. Using a novel human test, we identify limitations of current models in this task. In a final discussion, we examine how early-emerging concepts, incorporated in a hierarchy of concepts, may contribute to better learning. We propose two sources to this contribution – the discovery of useful fundamental concepts, and the role of early concepts in the decomposition of the learning task into simpler components. Finally, we consider the potential benefits of incorporating early human conceptual structure into large vision-language models. The main contributions of this work are as follows:
\begin{itemize}
\item Our study shows the computational advantage of systematically using a range of early acquired visual concepts in the learning of subsequent, more complex ones. 
\item We show that our human-like learning model contributes to the performance of standard models in terms of accuracy, efficiency and generalization. 
\item Using a new human experiment, we show limitations of current vision-language models in predicting human behaviour in a visual task. 
\item Based on our results, we propose a specific way of improving vision-language models by training their visual component on early human visual concepts. 
\end{itemize}

\section{Related Work}\label{RelatedWork}

\subsection{Infants' Understanding of Animacy and Goal Attribution}\label{RWInfants}
An early and influential contribution to understanding infants' goal encoding is Woodward's work \citep{woodward1998infants}, which demonstrated that infants distinguish between animate and inanimate entities in dynamic scenes, and attribute goals to agents or animate actors, but not to inanimate objects. This finding has been validated and elaborated through other experiments and variations \citep{woodward1999infants, woodward2000twelve, biro2011evidence}. A study by Luo and Baillargeon \citep{luo2005can} extended the understanding of goal attribution by showing that infants can attribute goals to new self-propelled objects. Further research illustrated that infants can generalize the attribution of goals to new actions when provided with sufficient contextual cues \citep{kiraly2003early}. In addition, infants exhibit a significant understanding of objects and their properties from an early age \citep{baillargeon1987object, spelke1995spatiotemporal, hespos2001reasoning}, expecting objects in the scene to follow physical rules, unlike animate agents that follow their goals \citep{spelke2022babies, lin2022infants}.

\subsection{Related vision models of infants early learning}\label{RWComputution} 
A large number of empirical studies have shown impressive capacities of young infants to understand different aspects of agent-object interactions in a dynamic scene, including the \textit{ownership of an object} \citep{friedman2008determining, saylor2011s}, \textit{Efficiency of agent-object interaction} \citep{gergely1995taking, phillips2005infants}, and \textit{using objects as tools} \citep{libertus2016infants}. In parallel to empirical studies of infant learning, computational models, including computer vision models, have been developed and used to replicate and better understand a range of related empirical findings.  

A recent study \citep{stojnic2023commonsense} investigated goal attribution in infants, both empirically and in vision models. The study found that infants anticipate agents' actions to be directed at objects rather than locations. In contrast, AI models often target locations, highlighting the need to integrate infants' understanding into models to better replicate human behavior. Li’s work \citep{li2024infant} introduced additional tasks related to agents and objects, along with a self-supervised model. This work demonstrated limitations of current neural network models in goal attribution tasks.

Several works \citep{bortoletto2024neural, hein2023comparing, zhi2022solving} have addressed infant-like tasks, using the Baby Intuitions Benchmark (BIB) \citep{gandhi2021baby}, a dataset specifically designed for evaluating developmental tasks related to agents. These studies employed various approaches, such as transformer-based architectures and Bayesian models, to tackle these tasks.

The main goal of the studies above was to develop models that replicate infant behavior. In contrast, the current work focuses on the fundamental differences between human learning and standard AI models. Specifically, we focus on the learning process itself: while infants demonstrate an early understanding of core concepts and progressively build on them, most AI models rely on end-to-end training without explicit decomposition of the hierarchical structure of learned concepts. By exploring this difference, we aim to highlight how integrating infant-like learning mechanisms could enhance the efficiency and generalization of AI systems.

\section{Method}\label{Meth}

In this section, we describe our approach to evaluating learning about goals and action prediction. Inspired by the empirical methods of Woodward et al. (1998), we created a new dataset to compare two models: cognitive and naive.

\textbf{The Cognitive Model} is akin to infant-like learning, where early-acquired concepts are integrated into later learning stages. In contrast, \textbf{the Naive Model} serves as a baseline, that does not use such concepts decomposition. This comparison allows us to explore and highlight the differences between these two learning approaches.

\subsection{Goal-Directed Dataset}
\begin{figure}[htbp]
\centering
\includegraphics[width=0.5\textwidth]{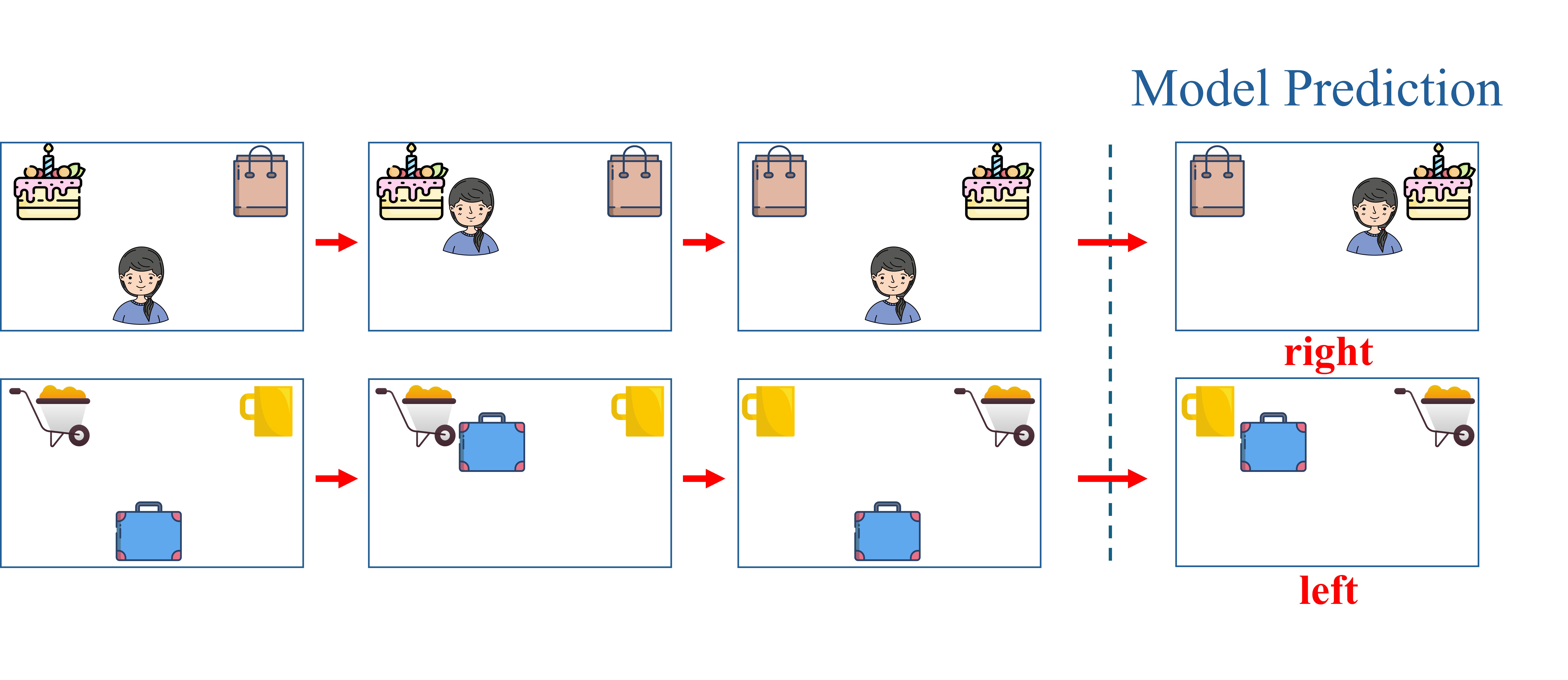}
\caption{Input data for Experiment 1: The model receives a sequence of three images and predicts the location ('left' or 'right') of the actor in the next unseen step of the sequence. Note that the objects switch locations in the third step. Top: animate actors; Bottom: inanimate actors. The person follows the object, the suitcase goes to the previous location.}
\label{fig1}
\end{figure}

We designed a dataset inspired by Woodward's experiments \cite{woodward1998infants}, extending it to include tasks beyond those studied in prior research. The dataset consists of sequences of simple scenes (Figure \ref{fig1}), created using icons from 'flaticon.com'. Each frame contains three entities: two inanimate objects and an actor, which can be either animate (e.g., hands, animals, people) or inanimate (e.g., books, flowers).

The primary task in most experiments was to predict the future motion of the actor in the next step of the sequence (which is unseen). For animate actors, following empirical findings, the prediction is determined by goals—meaning they are expected to move toward the same object they previously interacted with, even if its location has changed. In contrast, for non-animate actors, the prediction is determined by prior locations—meaning they are expected to repeat a prior trajectory, regardless of whether the object has changed.  Figure \ref{fig1} provides an example of the first task. This dataset allows us to test the ability of network models to predict goals and actions in scenes that involve both animate and inanimate entities.

\subsection{The Two Models}
\begin{figure}[htbp]
\centering
\includegraphics[width=\columnwidth]{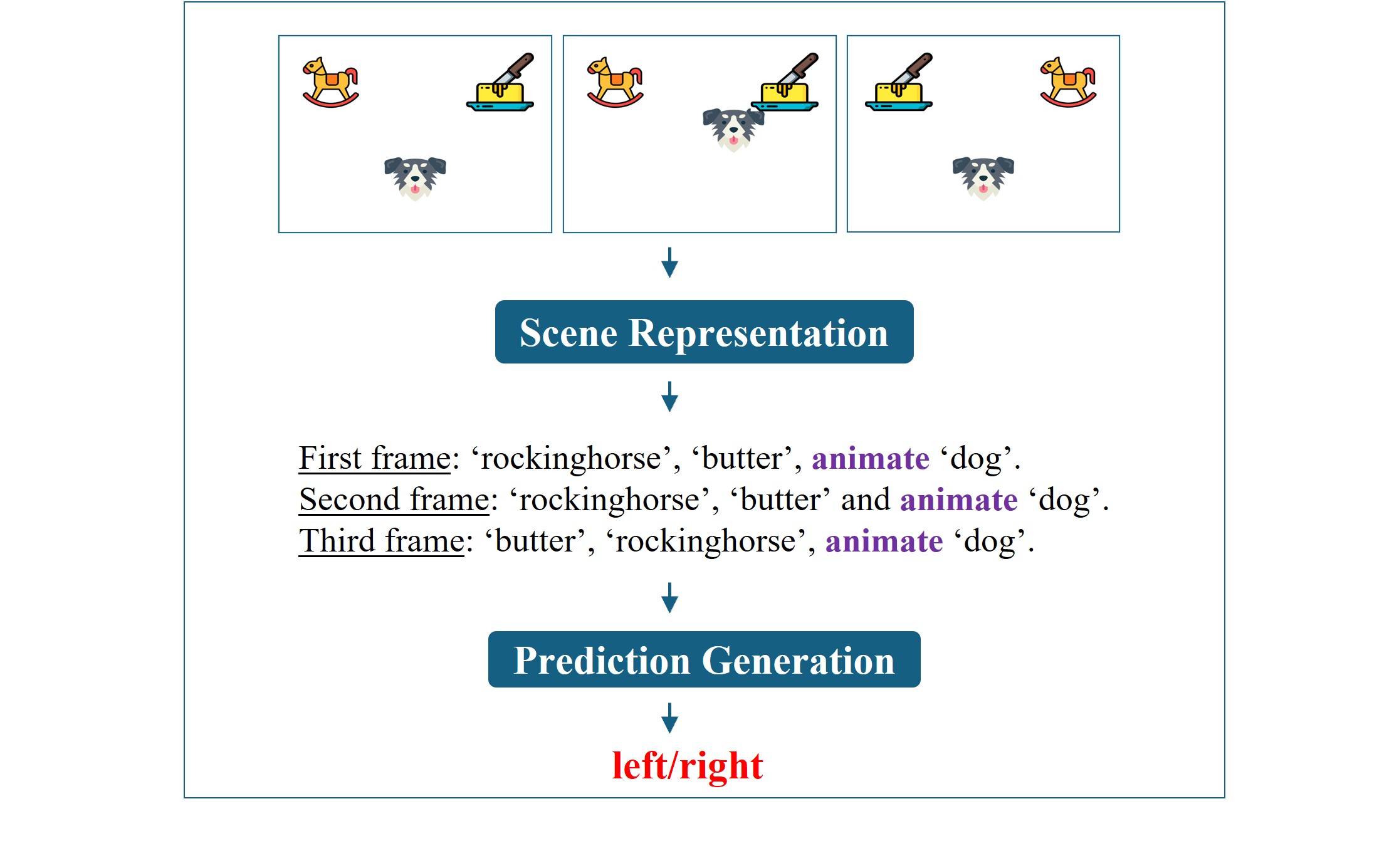}
\caption{Two-step process of the cognitive and naive models: First, scene representations are created, and then predictions are made. Concepts specific only to the cognitive model are shown in bold and purple.}
\label{fig2}
\end{figure}

We compared two models to evaluate the different types of learning. Both models follow the same two-step process: (Figure \ref{fig2}) First, frames are processed to create visual representations of the scene, encoding the class and location of entities. Then, using these scene representations, the models predict the future location of the actor ('left' or 'right'). 

The key difference between the models lies in their treatment of conceptual information: Following results of Woodward-like experiments, in which infant subjects already differentiate between animate and non-animate actors, the \textbf{Cognitive Model} incorporates additional concepts, e.g. distinguishing between animate and inanimate actors, into its representation, whereas the \textbf{Naive Model} relies solely on raw scene representations, excluding such conceptual information. Below, we elaborate on each of the steps in more detail.

\subsubsection{First Step - Scene Representations}
The objective of the first step is to create scene representations for the image sequences. To achieve this, we fine-tuned the BLIP model \cite{li2022blip} to generate representations for each frame, encoding the class and location of the three entities (upper-left, upper-right, bottom). For example, the representation for the first frame in the top row of Figure \ref{fig1} would be: 'cake', 'bag', 'girl'. The meaning of the words (e.g., 'cake', 'bag', 'girl') is not used by the model. Instead, the words only serve as placeholders to encode the entities' identities.

In the cognitive model, the actor type (animate/inanimate) is explicitly included during this step. For example, the cognitive model's representation for the previous scene would be: 'cake', 'bag', 'animate girl'.

\subsubsection{Second Step - Prediction Generation}
 The prediction task is framed as a classification problem using the BERT model \cite{devlin2018bert}. The input to the model is the full sequence representation of the frames, and the model's output is a prediction of the actor's future location ('left' or 'right') in the next step of the sequence. The results in the following sections evaluate the model's accuracy in this prediction. Training and optimization details are provided in Appendix~\ref{app_technical}.
 
We used large models (BLIP, BERT) in our cognitive model to reflect the complexity of the infant brain, as infants, too, are born with a sophisticated neural architecture. We fine-tuned all layers in both models for our specific tasks. To ensure that the pre-trained linguistic knowledge of the models did not drive the observed effects, we also ran a version of the experiment using binary representations (e.g., 0/1 vectors) instead of words in natural language and obtained similar results. Furthermore, we confirmed that their prior knowledge did not bias the results by verifying that the initial performance of the models in our tasks was not significantly different from the chance level. 

\section{Experiments}\label{Exp}

\subsection{Experiment 1: Prediction}

Experiment 1 used the three-frame paradigm shown in Figure \ref{fig1}, where the task was to predict the actor’s location (left vs. right) following three frames. We compared how the cognitive and naive models learned this task. The only difference between the models is the inclusion of the animate/non-animate label in the cognitive model. 

The primary question was how the two models compare in learning the prediction task. Two important points regarding this comparison are:
1. The naive model could theoretically learn to distinguish between the two types of actors based on the data it observes, as the actor type is consistently correlated with its behavior in both training and test scenarios.
2. Note that the addition, such as the labels ‘animate’ and ‘non-animate’, is provided in a 'bare' form, without additional context about its implications or how it relates to the model's predictions. For the model, these labels are arbitrary markers with no associated information.

\subsubsection{Data}

The models were trained using sequences of three frames (Figure \ref{fig1}), employing various combinations of icons, with half of the sequences showing animate actors and half showing non-animate actors. Two dataset sizes were used: a small dataset (320 training examples and 80 test examples) and a large dataset (four times larger, 1280 training, 320 test examples). The test data used the same actors as the training data, but introduced novel target objects to evaluate generalization. Both the cognitive and naive networks were trained for 2000 epochs. Each experiment was repeated 15 times with randomly generated sequences, and the results were averaged.

\subsubsection{Results}

\begin{figure}[htbp]
\centering
\includegraphics[width=\columnwidth]{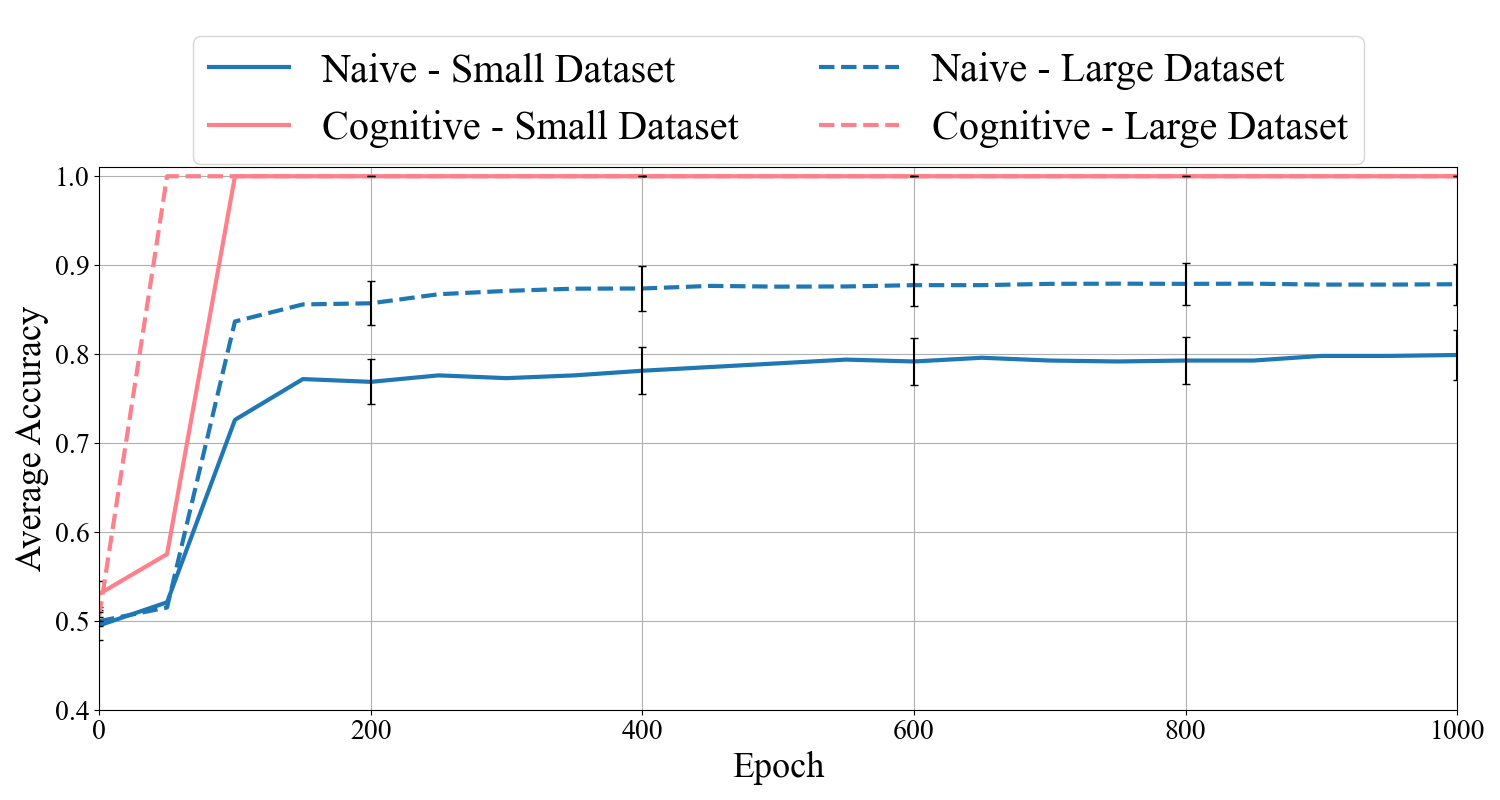}  
\caption{Experiment 1: Prediction Results. Average test accuracy of the naive and cognitive networks for both the small and large datasets, with the Standard Error of the Mean (SEM) included.} 
\label{three frames}
\end{figure}

The averaged results of Experiment 1 for both models are shown in Figure \ref{three frames}, comparing their performance on the small and large datasets. Only the first 1000 epochs are shown, as the results stabilized within this range. The cognitive network achieved perfect accuracy after 100 epochs with the smaller dataset, and converged to perfect accuracy after 50 epochs with the larger dataset. In contrast, the naive network achieved approximately 80\% accuracy after 1000 epochs with the smaller dataset and 87\% accuracy with the larger dataset. 

These results demonstrate that the cognitive network consistently achieves higher accuracy (perfect performance) and faster convergence, even with a small dataset.

\subsection{Experiment 2: Generalization}
Generalization to new tasks and domains is a crucial aspect and a useful measure of intelligent behavior \cite{lake2018generalization, chollet2019measure}. In the previous experiment, the test data used the same actors as in the training but introduced new target objects. Here, we extended this to assess the networks' ability to generalize by evaluating their performance with entirely new actors and a novel spatial configuration.

\begin{figure}[htbp]
\centering
\includegraphics[width=\columnwidth]{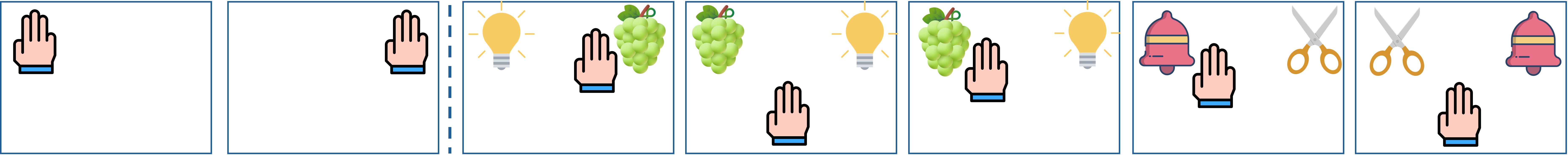}
\caption{Input Data for Experiments 2 and 3: The five rightmost frames are used in Experiment 2, while the full seven  frames are used in Experiment 3.}
\label{5-7-data}
\end{figure}

To address generalization, a 5-frame paradigm was introduced (Figure \ref{5-7-data}). This paradigm first presents the actor's behavior in the initial three frames and then introduces the same actor with new target objects, requiring the model to predict the actor's behavior in the next step. Unlike Experiment 1, this setup includes information about the actor's behavior within the data itself, theoretically enabling both networks to predict an actor's behavior from a single example. As a result, it is well-suited for testing generalization to new actors.

\subsubsection{Experiment 2.1: Generalization to New Actors}

\textbf{Data.} The 5-frame paradigm incorporates the second and third frames from Experiment 1 (Prediction task), along with the frame that features the next step of the sequence, which was not shown in Experiment 1. The first three frames of Experiment 2 already show the actor's preference. This sequence is extended by adding two new frames that present the same actor with new target objects. The prediction task focuses on the actor's behavior in relation to these new targets. The models were trained using sequences of five frames, each featuring various combinations of icons, with half of the sequences containing animate actors and the other half containing non-animate actors.

Two task types were generated: Task 1 (T1): A simpler task, where the test data uses the same actors as in the training data but introduces new target objects. Task 2 (T2): A more challenging task because the test data includes entirely new actors and new target objects. 

The networks were first trained on T1 and then further fine-tuned under one of two conditions: T1-T1: The same simple task (T1) with additional data. This condition served as a control to assess the effect of additional data without introducing new challenges.
T1-T2: Following the initial task T1, the model is fine-tuned for the more challenging generalization task (T2). Each task used 640 training examples and 160 test examples. Both networks were trained for 1000 epochs. Each experiment was repeated 15 times with randomly generated sequences, and the results were averaged.
 
\begin{figure}[htbp]
\centering
\includegraphics[width=\columnwidth]{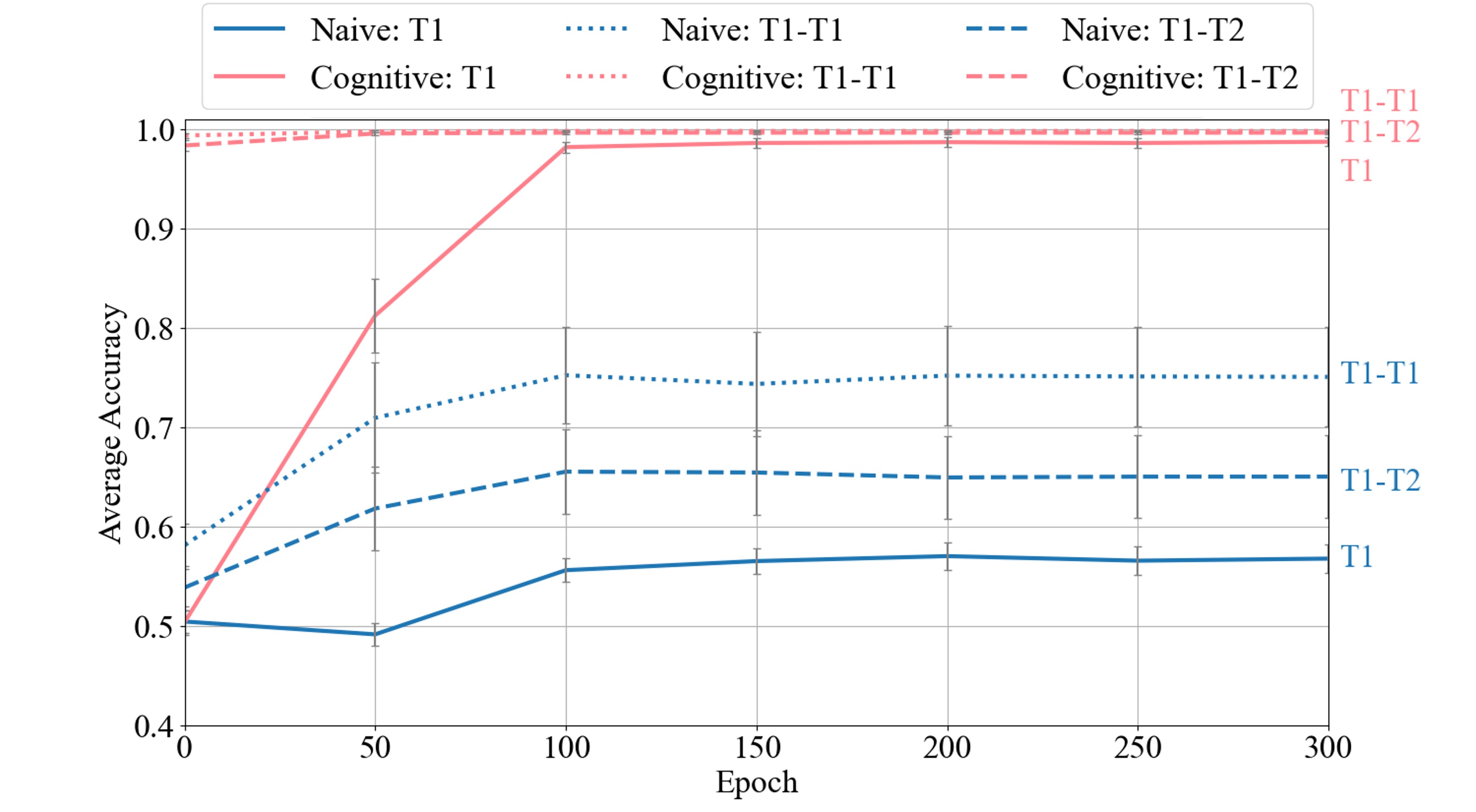}
\caption{Experiment 2.1: Generalization to New Actors Results. Average test accuracy of the naive and cognitive networks on Task 1 (T1), retraining on Task 1 (T1-T1), and retraining on Task 2 (T1-T2), with SEM included.}
\label{five frames}
\end{figure}

\textbf{Results.} The results for both models are shown in Figure \ref{five frames}, comparing the performance under the following conditions:
1. Learning the initial task (T1).  
2. Retraining on the same task with additional data (T1-T1).  
3. Generalization to a new task (T1-T2). Only the first 300 epochs are shown, as there was no significant change afterward. The cognitive network outperformed the naive network in both accuracy and learning speed across all tasks: T1-T1 Condition: The cognitive network achieved near-perfect accuracy in the first run and 100\% accuracy almost immediately in the second run. The naive network stabilized at 58\% accuracy after 1000 epochs in the first run and reached 75\% accuracy after 200 epochs in the second run. T1-T2 Condition: For the challenging generalization task, the cognitive network achieved perfect accuracy after just a few epochs. The naive network, however, reached only ~65\% accuracy, performing worse than in the T1-T1 condition.

We conclude that in addition to a difference in final performance, there is also a marked difference in the ability to generalize to a somewhat different, more challenging task.

\subsubsection{Experiment 2.2: Spatial Generalization}

\textbf{Data.} We used the same five-frame paradigm described above for the spatial generalization experiment. As in the previous section, both networks were first trained on Task 1 (T1). We then introduced two spatial generalization conditions. In the first condition (T1-Generalization), the networks were initially trained on T1 and then retrained on a spatially modified version of the task obtained by horizontally flipping the original five-frame configurations. For example, objects that previously appeared at the top of the frame were relocated to the bottom. The actors remained the same as in T1. In the second condition (T1-Spatial+Actors), the networks were trained on T1 and then retrained on the horizontally flipped configurations with entirely new actors. Each task used 640 training examples and 160 test examples. Both networks were trained for 300 epochs. Each experiment was repeated 15 times with randomly generated sequences, and the results were averaged.

\begin{figure}[htbp]
\centering
\includegraphics[width=\columnwidth]{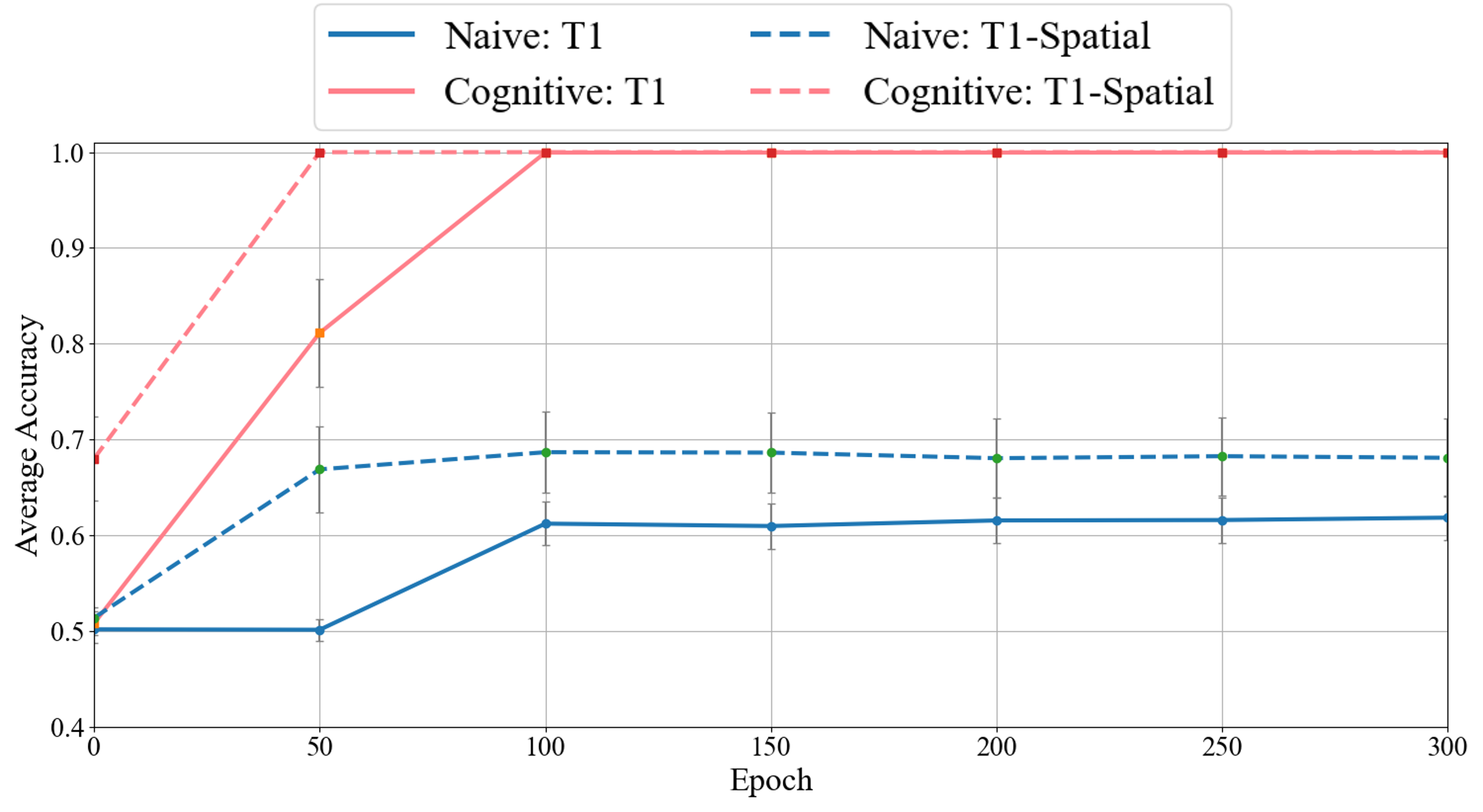}
\caption{Experiment 2.2: Spatial Generalization Results. Average test accuracy of the naive and cognitive networks on Task 1 (T1) and retraining on Spatial task (T1-Spatial), with SEM included.}
\label{spatial}
\end{figure}

\begin{figure}[htbp]
\centering
\includegraphics[width=\columnwidth]{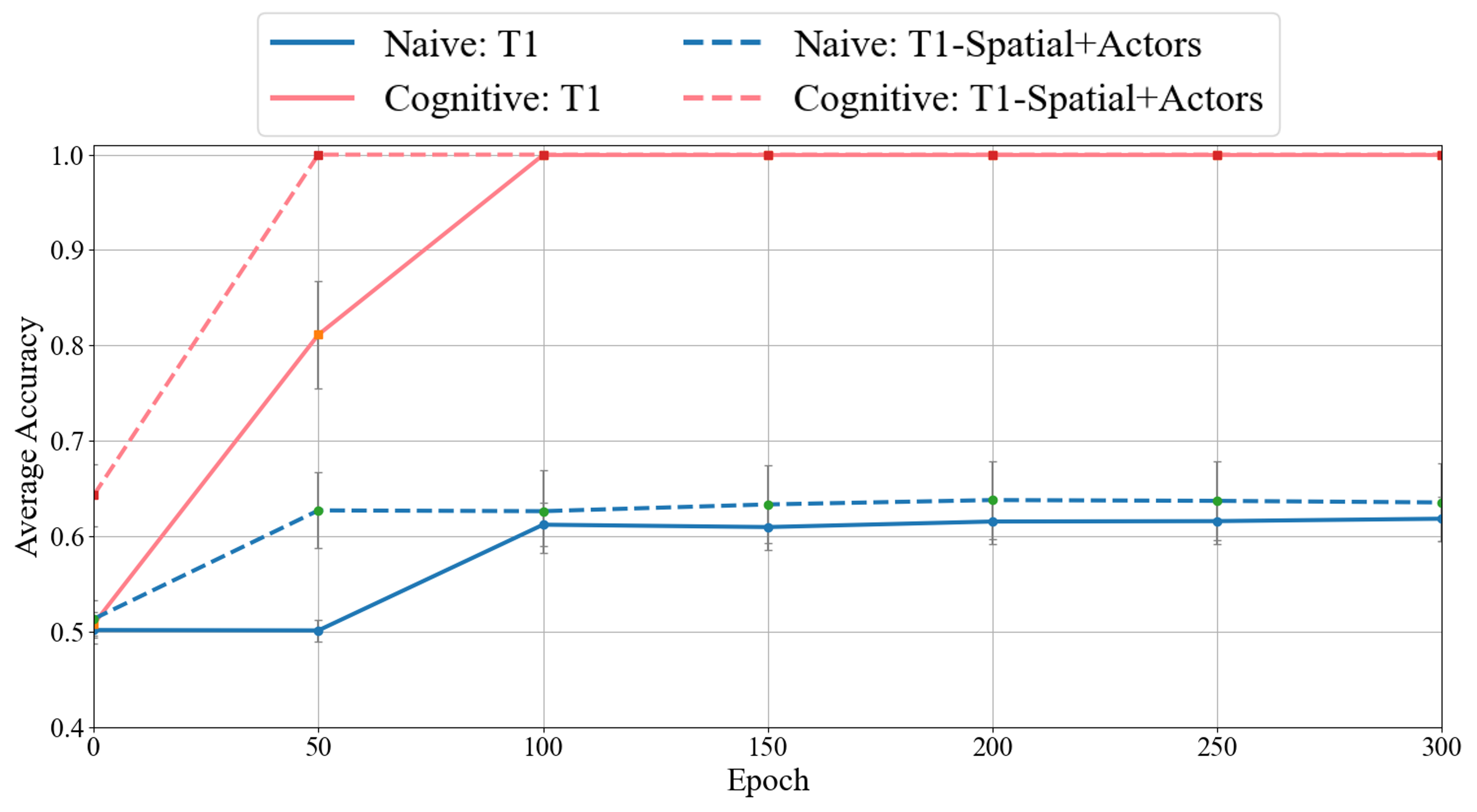}
\caption{Experiment 2.2: Spatial Generalization with New Actors Results. Average test accuracy of the naive and cognitive networks on Task 1 (T1) and retraining on Spatial and New Actors task (T1-Spatial+Actors), with SEM included.}
\label{spatial challenging}
\end{figure}

\textbf{Results.}
The results for both models are shown in Figures \ref{spatial} and \ref{spatial challenging} for the two spatial generalization conditions, respectively. The initial T1 training phase is identical across both conditions (for clarity, we display the results from one representative run) and corresponds to the same T1 task used in Experiment 2.1. Minor differences in accuracy reflect natural variation across runs, as captured by the SEM. In this setting, the naive network reached accuracy of 61\% after 100 epochs, while the cognitive network achieved near-perfect accuracy within the same number of epochs. In both spatial generalization conditions, the cognitive network outperformed the naive network in terms of both final accuracy and learning speed. In the T1-Generalization condition (horizontal spatial flip with the same actors), the cognitive network reached near-perfect accuracy on the second task within approximately 50 epochs. The naive network, in contrast, reached only 68\% accuracy after 100 epochs and showed no substantial improvement thereafter. In the more challenging T1-Generalization+Actors condition (horizontal flip combined with new actors), the cognitive network again achieved near-perfect accuracy rapidly. The naive network reached approximately 63\% accuracy after 150 epochs and maintained this level without further improvement. 

Importantly, in both spatial generalization settings, the cognitive network began the second task with an initial accuracy higher than that at the beginning of the original T1 training. This suggests that knowledge acquired during T1 generalized to the modified tasks. In contrast, the naive network showed little evidence of such generalization: its initial performance on the spatially modified tasks was similar to its starting point in the original T1 phase, indicating minimal generalization from prior learning.

These results demonstrate that the difference between the cognitive and naive networks is not limited to final performance levels but also reflects differences in generalization abilities. The cognitive network not only achieves higher final accuracy, but also adapts more rapidly to spatially modified and actor-novel settings.

\subsection{Experiment 3: Decomposition} 
We address the following question: How is the additional information provided to the cognitive model learned? In earlier experiments, the actor's type (animate/inanimate) was explicitly provided to the cognitive network. In the current experiment, however, the same information is available to both models. Inspired by evidence that self-propelled motion is a strong indicator of animacy in infants \cite{luo2005can}, the experiment introduced additional frames to enable the learning of this concept.

The cognitive network emulates infant learning by first acquiring simple concepts (e.g., animacy) and then integrating this knowledge into downstream tasks. In contrast, the naive network processes the entire dataset directly, without explicitly decomposing the task.
 
\subsubsection{Data}
The experiment builds on the previous 5-frame paradigm (Figure \ref{5-7-data}) by adding two additional frames at the beginning of each sequence. These new frames depict the actor's movement: If the actor changes location between the first and second frames, it is classified as self-propelled (indicating animacy). If no change in position is observed, the actor is classified as non-self-propelled (indicating inanimacy). The models were trained using sequences of seven frames, each featuring various combinations of icons, with half of the sequences containing animate actors and the other half containing non-animate actors.

For the cognitive model, training first focuses on learning the concept of animacy (animate vs. inanimate), which is associated with the actor and subsequently used in the five-frame task. In contrast, the naive network receives all seven frames concatenated and learns the prediction task directly, without explicit concept decomposition. 

The naive network was tested on the 7-frame paradigm with progressively larger datasets, starting from 640 training and 160 test examples, doubling in size at each step. The cognitive network, in contrast, learned actor types using smaller datasets of 320 training and 80 test examples, which proved sufficient. For the 5-frame task, a dataset of 640 training and 160 test examples was used. Each experiment was repeated 15 times with randomly generated sequences, and the results were averaged.  Both networks were trained for 300 epochs due to the larger datasets.

\subsubsection{Results}
\begin{figure}[htbp]
\centering
\includegraphics[width=0.85\columnwidth]{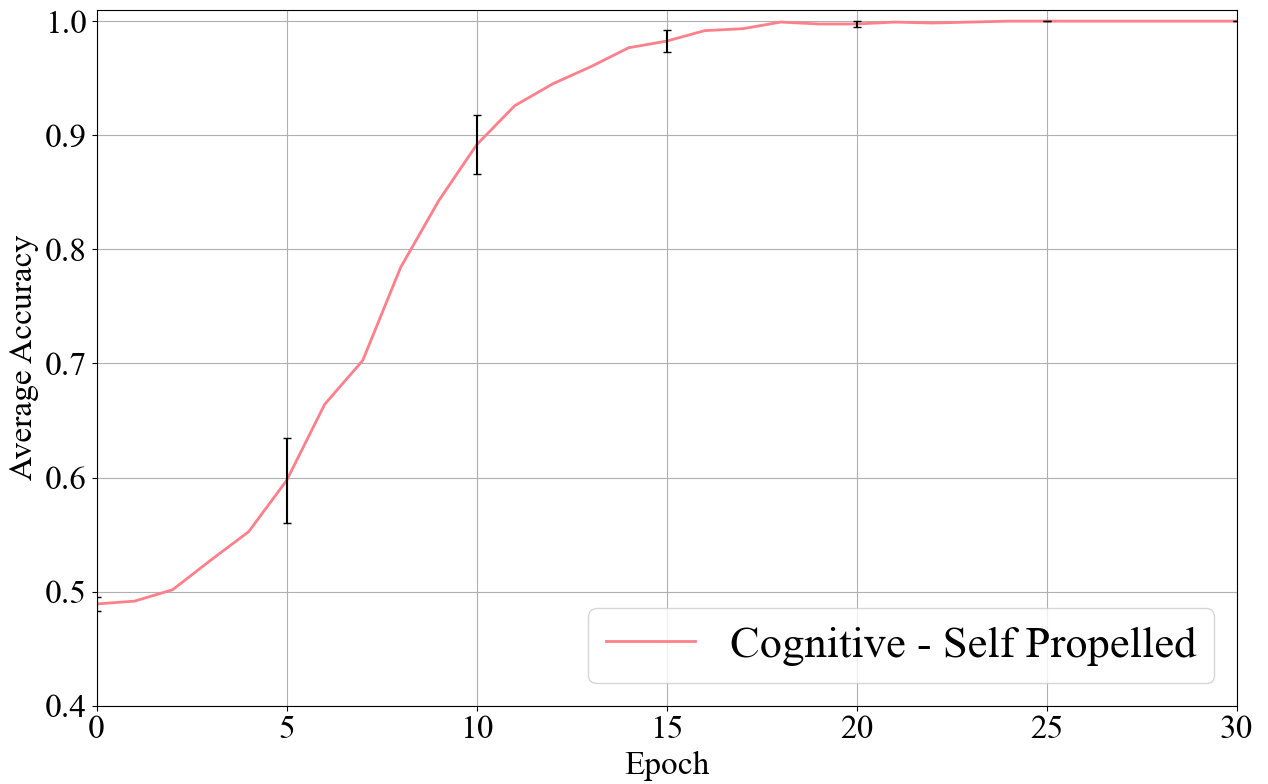}
\caption{The cognitive network's performance in the preceding step, where it learned to distinguish between animate and inanimate entities based on self-propulsion.}
\label{self}
\end{figure}

\begin{figure}[htbp]
\centering
\includegraphics[width=\columnwidth]{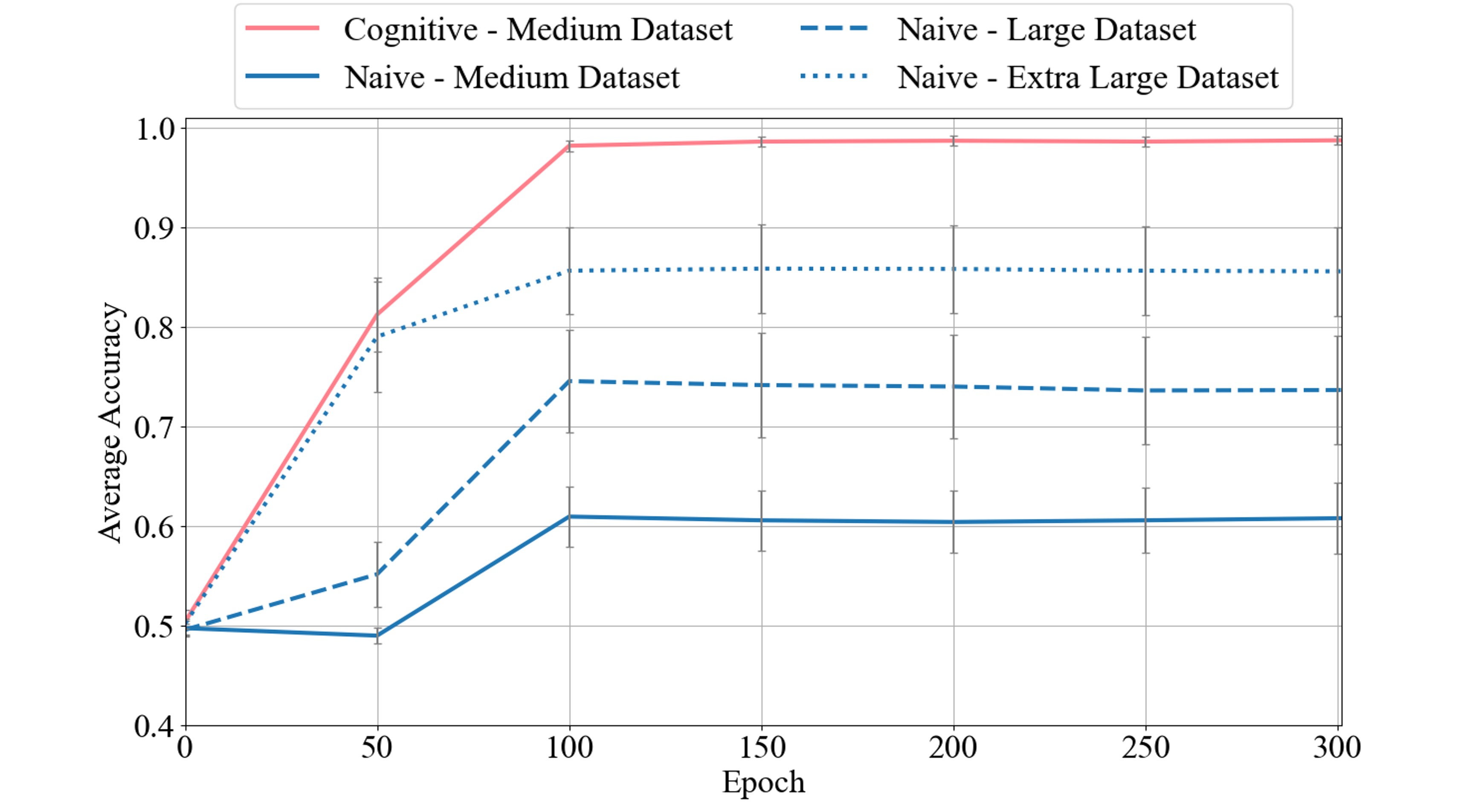}
\caption{Experiment 3: Decomposition Results. Average test accuracy of the naive network trained with varying dataset sizes, compared to the cognitive model's performance on the 5-frame test with medium dataset size, with SEM included.}
\label{seven frames}
\end{figure}

The cognitive network learned the first task of classifying the actor's type within 20 epochs, achieving perfect accuracy (Figure \ref{self}). For the subsequent task, the cognitive network performs the 5-frame paradigm discussed earlier, so the results for this stage are identical to those of the 5-frame test. The performance of the naive network is shown in Figure \ref{seven frames}, alongside the cognitive network's results from the 5-frame stage.  For the smallest dataset, the naive network achieved approximately 60\% accuracy. When the dataset size was doubled, its accuracy increased to 74\%. With the largest dataset, accuracy improved to 85\%. In comparison, the cognitive network achieved near-perfect accuracy on the same 5-frame test, even with the smallest dataset. 

The results show that the decomposition of the task into subtasks, performed by the cognitive network, leads to substantial improvement in performance.  

\subsection{Experiment 4: Goal Attribution}
In this experiment, a related yet distinct concept, 'Goal,' is investigated. Similar to the idea of actor type, it is assumed, based on the empirical literature, that infants can attribute goals to animate entities, based on cues such as certain types of contact or gaze direction \citep{woodward1998infants, phillips2002infants}. Similar to the 'animacy' concept, the concept of 'having a goal' is provided to the cognitive model, and the results are compared with a naive model that can infer the goal, but is not directly provided with the concept.

\subsubsection{Data}  
The Goal Attribution task uses a one-frame paradigm, where each frame depicts an actor with two target objects, similar to the first frame of the three-frame paradigm introduced earlier. Unlike previous experiments, all actors in this task are animate, as goal attribution applies only to animate entities. Each training dataset includes 10 repetitions of an actor appearing with the same goal object, but with varying distractor objects across frames. The data demonstrates that the actor has a global goal, consistently preferred over all distractors. During testing, the actors and their associated goals remain the same as in training, but the distractors are new.

For the cognitive model, the goal is provided as part of the actor's representation (e.g.: 'girl with goal cake'), but its implications are not specified and must be learned during training. In contrast, the naive model must infer the goal directly from the data. The only difference between the models is the inclusion of the goal representation in the cognitive model. An example of the goal attribution representation and training data can be found in Appendix~\ref{app_goal}.

The training size was 320, and the testing size was 80. Both models were also evaluated on their ability to generalize to a more challenging task (T2), involving new actors and new goals. The experiment ran for 1000 epochs, but only results from the first 100 epochs are presented, as there was no significant change beyond that point. Each experiment was repeated 15 times with randomly generated sequences, and the results were averaged. 

\subsubsection{Results}
\begin{figure}[htbp]
\centering
\includegraphics[width=\columnwidth]{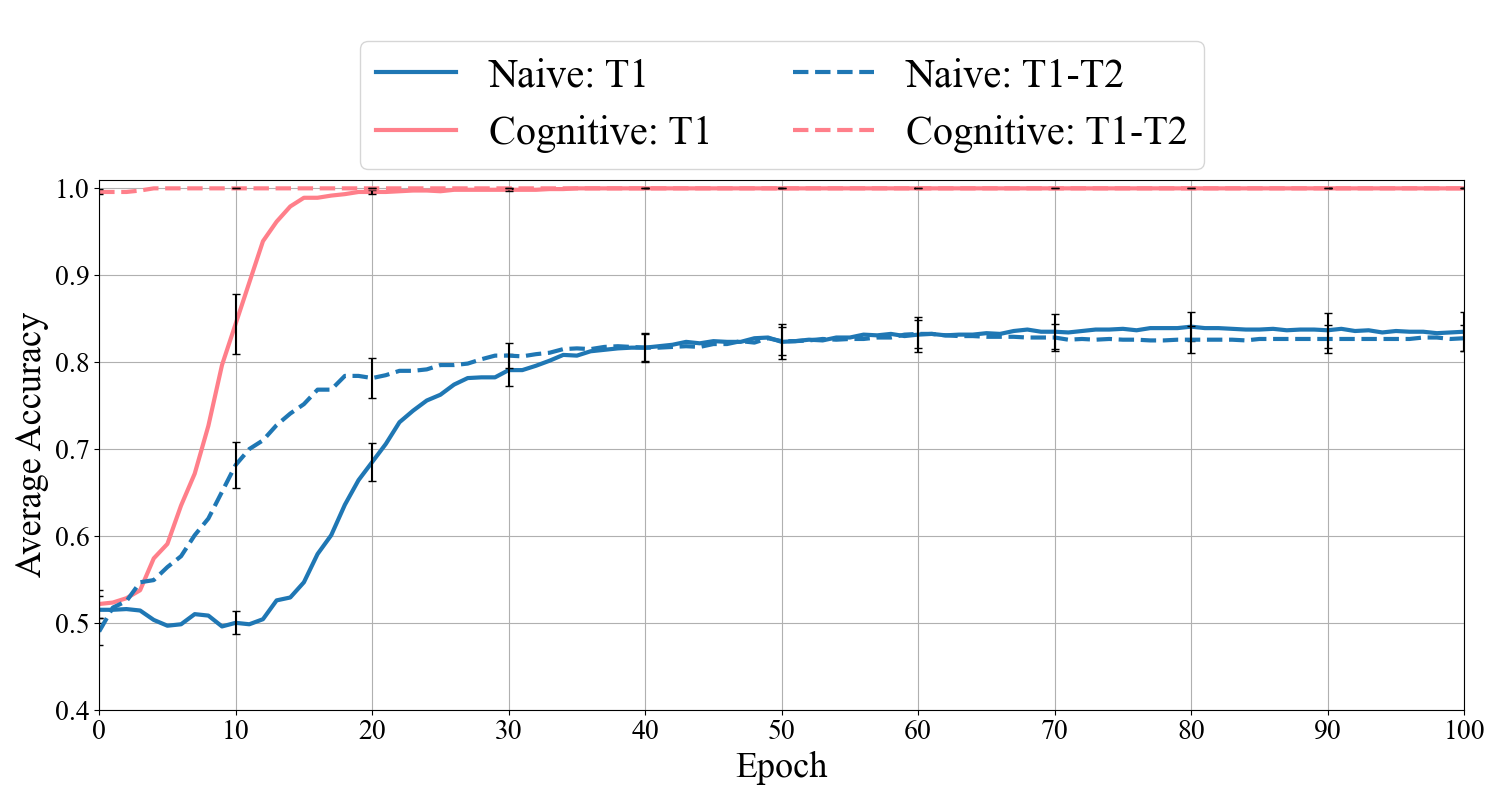}
\caption{Experiment 4: Goal Attribution Results. Average test accuracy of the cognitive and naive networks on the first and second tasks, with SEM included.}

\label{global goal}
\end{figure}

The results for the Goal Attribution task are shown in Figure \ref{global goal}. The cognitive network achieved perfect accuracy after around 20 epochs on the first task (T1) and maintained perfect accuracy almost immediately in the generalization task (T2). The naive network, in contrast, reached approximately 82\% accuracy for both tasks after 100 epochs. Although the learning rate was faster for T2 than T1, the overall accuracy of the naive network did not improve significantly.

Generalization behavior is assessed by examining the transition from T1 to T2. In the cognitive model, there is full transfer: after learning T1, the accuracy of T2 (dashed curve) starts at the same level that T1 reached at the end of training. In contrast, the naive model shows limited transfer. The accuracy of T2 begins at chance level, similar to T1, but exhibits an accelerated learning rate, indicating partial transfer of knowledge from the first task to the second.  These results demonstrate that the cognitive model learns faster, achieves higher accuracy, and transfers knowledge seamlessly to the generalization task (T2). In contrast, the naive model learns less efficiently and struggles with knowledge transfer.

\section{Do Large Vision–Language Models Exhibit Human-Like Behavior in the Woodward Task?}

Learning in humans and in our cognitive model begins with simple concepts that guide prediction and shape the final representations. In contrast, current Vision–Language models (VLMs) and Multi-Modal Large Language Models (MMLLMs) are typically trained end-to-end with the goal of optimizing performance in a final task. In this section, we examine whether VLMs acquire the core concepts that are part of the human cognitive system as part of their training. 

Specifically, we examined whether the behavior of VLMs resembles human behavior in the Woodward task. In humans, predictions in the Woodward configuration are consistent with the notion that animate agents move toward their preferred object targets, whereas inanimate objects do not show such a preference, but show some preference to repeat their previous trajectories. We tested whether VLMs exhibit similar predictive behavior, and our evaluation focused on the performance of VLMs following their general training, without specific fine-tuning for the Woodward task.

 Testing VLMs on the Woodward-type tasks is a part of a broader comparison, since object-tracking and trajectory-tracking behaviors, as studied in the Woodward task, are instances of a more general and fundamental distinction between the behavior of animate versus inanimate entities. The behavior of animate entities is determined by their internal goals, preferences, and beliefs, whereas non-animate behavior is determined by external forces and circumstances. For humans, this is an intuitive and basic dichotomy between ‘intuitive physics’ and ‘intuitive psychology’ \citep{lake2017building}, which begins to emerge already at early development \citep{spelke2022babies}. We describe below our comparison of human subjects and VLM models in a version of the Woodward task. As described in detail below, we compare their capacity to correctly predict the behavior of animate and non-animate agents in the Woodward test scenario, with the main focus on predicting that animate agents prefer object-tracking (preference for moving towards the preferred object) over trajectory-tracking (preference to repeat a previous trajectory). 

\subsection{Human testing}

\begin{figure*}[t]
\centering
\includegraphics[width=\textwidth]{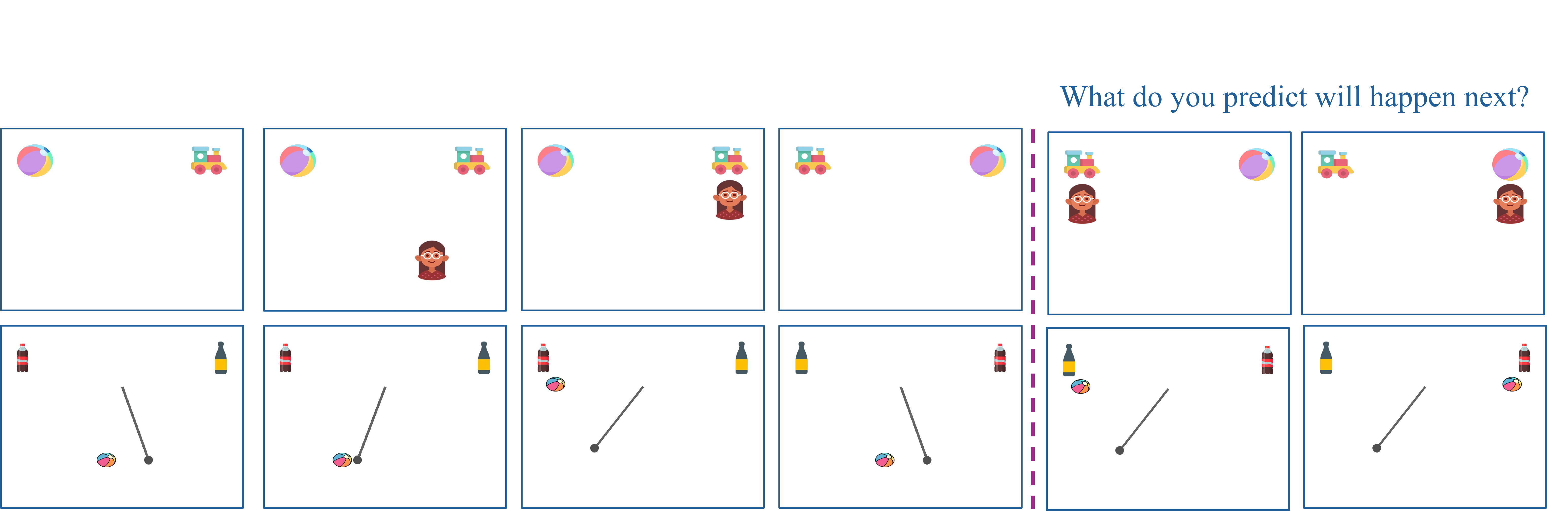}
\caption{Example data from the human experiment and VLM testing. Top: Animate condition. Bottom: Non-animate condition. Left: Example frames from the videos presented to human participants and models. Right: An example of the question presented after the video.}
\label{fig_vlm}
\end{figure*}

\subsubsection{Participants}
We recruited 200 participants through the Prolific research platform. All participants provided their informed consent and were compensated for their participation. Participants completed the experiment online.

\subsubsection{Method}  
To test human behavior in Woodward-like tasks, we designed two types of video conditions: animate and non-animate (see Figure \ref{fig_vlm}). The dataset included 100 animate videos and 100 inanimate videos. Each participant viewed a single video.

In the animate condition, each video begins with two goal objects located at the top of the screen. An animate actor (e.g., a girl or a dog) then enters from the bottom of the screen and moves toward one of the objects until reaching it. A gray screen is then shown, after which the two objects reappear with their locations swapped. After the video ends, participants are shown two images and asked to predict which one is more likely to be the next part of the video. In one image, the actor reaches toward the same goal object, which is now in a new location. In the other image, the actor reaches toward the same location as before, which now contains the other object.

In the non-animate condition, the videos show two objects at the top of the screen and a non-animate actor (e.g., a ball) at the bottom. Next to the non-animate actor there is a pendulum, which provides the physical cause of the motion by pushing the actor and causing it to move in one direction on the screen and stop near one of the objects. A gray screen is then shown, and the two objects reappear with their locations swapped. The prediction task is similar to the animate condition. The participants see two images, showing the actor at the bottom near the pendulum and the two objects at the top. In each image, the non-animate actor is positioned near one of the sides, and participants must choose which image is more likely to represent the continuation of the video. Participants were also asked to provide an explanation for their prediction after choosing one of the images. See Appendix~\ref{app_human_exp} for additional details about the human experiment.

\subsubsection{Results}
Out of 100 participants in the animate condition, 75 answered correctly, and 81 out of 100 participants answered correctly in the non-animate condition. Unlike the original Woodward experiment, which was conducted with infants, our experiment was conducted with adults. The results in the animate condition are consistent with the findings of the original study. In contrast, in the original Woodward experiments the non-animate condition was close to chance level, and this preference develops only later.

\subsubsection{VLM Testing}

\begin{table*}[t]

\begin{tabular}{lccccc}
\toprule
& Humans & Gemini 3.1 Flash & Gemini 3.1 Pro &  Claude 3.5 Sonnet & Claude Opus 4.6  \\
\midrule
Animate & 75 & 41$^{*}$& 63 & 53$^{*}$ &37$^{*}$  \\
Non-animate & 81 &47$^{*}$  & 90 & 50$^{*}$ & 94$^{*}$ \\
\botrule
\end{tabular}

\vspace{6pt}

\begin{tabular}{lccccc}
\toprule
 & Humans & GPT-4o & GPT-5.2 & Qwen VL Max & Qwen 3.5 Plus \\
\midrule
Animate & 75 & 35$^{*}$ & 84 & 48$^{*}$& 85  \\
Non-animate & 81 & 60$^{*}$ & 58$^{*}$ & 49$^{*}$ & 88 \\
\botrule
\end{tabular}

\caption{Accuracy of humans and vision-language models on the animate and non-animate conditions (N = 100 trials per condition). Models marked with $^{*}$ were significantly different from human performance (two-proportion z-test, $\alpha = 0.05$).}
\label{tab:vlm_results}

\end{table*}
\subsubsection{Models} 
To evaluate model behavior in Woodward-like tasks, we benchmarked eight models: two Gemini models, two GPT models, two Qwen models, and two Claude models. For each model family, we selected two versions: the most recent available model and an earlier-generation version. 

\subsubsection{Method}
The VLM test was similar to the human experiment and used the same data. The models were presented with the videos and prompted to predict the next step by choosing between two possible images (A or B). This procedure was applied to both the animate and non-animate conditions. The models were instructed to explicitly provide a prediction (A/B) along with an explanation. Additional details and statistical analysis details are provided in Appendix~\ref{app_vlm_details}

\subsubsection{Results}
 We compared the models’ predictions with the ground truth for both conditions and across all models, see table \ref{tab:vlm_results}. In general, the more recent model versions performed better than the earlier-generation models in both conditions. To assess alignment with human performance, we conducted two-proportion z-tests comparing each model to humans within each condition ($\alpha = 0.05$). In the animate condition, most models performed significantly lower than humans. Gemini 3.1 Pro was lower than humans but did not differ significantly. GPT-5.2 and Qwen 3.5 Plus achieved higher accuracy than humans; however, these differences were not statistically significant. In the non-animate condition, most models again performed significantly lower than humans. Gemini 3.1 Pro and Qwen 3.5 Plus achieved higher accuracy than humans, but the differences were not statistically significant. Claude Opus 4.6 was the only model that performed significantly higher than humans in this condition.
 
 The difference between the human system and the VLMs in the tasks described above is striking if we take into account the amount of data and training used in the learning process. The training of advanced LLMs requires a huge text data set, measured in up to trillions of text tokens (e.g. 15 trillions in Meta’s Llama) and billions of image-text pairs (e.g. 5.86 billions in LAION-5B). It is also notable that 6–9-month-old infants’ performance in the animate condition (approximately 75\%; \citep{woodward1998infants}) in a similar task is relatively high compared with the models we tested.

\section{Discussion}\label{Disc}

The reported results of the current studies demonstrate significant effects of early cognitive concepts, leading to efficient learning and broad generalization. A natural question that arises is whether similar benefits would extend to more natural real-world applications, and to the training of current large models. In this section, we briefly discuss two fundamental issues related to the results and their implications. First, we discuss how early-emerging concepts, incorporated in a hierarchy of concepts, may contribute to better learning. Second, we consider the possibility of incorporating the equivalent of early human concepts into current vision and vision-language models. 

\subsection{Possible Contributions of the Early Cognitive Concepts to Better Learning}

Two main factors appear to be the basis for the advantage of the cognitive models.   
The first has to do with the discovery and incorporation of useful and informative concepts. When we learn about a new domain, we often benefit from getting acquainted with concepts that capture basic entities and interactions in the domain; e.g., concepts such as supply, demand, equilibrium, incentives, and competition are useful for understanding aspects of the economy. The suggestion is that the discovery of useful concepts and their interactions can be a demanding part of the learning process; therefore, having them available can lead to more efficient learning and a better understanding of the domain. The concepts of 'Animacy' and 'Goal' discussed in the studies above were shown to be useful in capturing a basic distinction between two major categories of entities:  animated agent vs. inanimate objects. The general conjecture is that the conceptual structure that emerges in early human visual learning, selected by evolutionary processes, can be useful in the learning and understanding of relevant domains. 
 
A second benefit of early cognitive concepts for learning a given task may also come from their role in decomposing the learning of the task into simpler components. There is empirical and theoretical evidence suggesting that the difficulty of learning a function \textit{F} by a deep network model depends on the complexity of \textit{F}. As complexity increases, learning is likely to require more data, and the probability of identifying \textit{F}, or a close approximation, starting from a random initialization, is likely to decrease. Empirical support for these relations was shown for several complexity measures, e.g. in \citep{valle2018deep}. If this notion is correct, the training of a model may benefit from a decomposition of the target function into components, where the complexity of each component is lower than the complexity of the full task. This suggests that the cognitive model used in our study and similar models could benefit from a hierarchical decomposition, in which simple early concepts are learned on their own first and then used to support the learning of more complex concepts.

\subsection{Possible Implications for Current Vision and Vision-Language Models}\label{DiscVision}
Our study has demonstrated that incorporating early cognitive structures can enhance the learning process, making it more efficient and accurate, and can lead to broader generalization, thereby facilitating a deeper understanding of the learned task. An important open question is whether the benefits shown in a simple example could also apply to large vision and vision-language models, and to complex real-world tasks. Current large models have achieved impressive performance, but they are still far from perfect. For example, models have been shown to sometimes ‘show failures in surprisingly trivial problems’ \citep{dziri2024faith} or ‘show basic errors in understanding that would not be expected even in non-expert humans’ \citep{oh2024generative}. 

 One common approach to improve existing models is to increase and improve the data used for training. A complementary approach may be based on the incorporation of useful early concepts within a hierarchical conceptual structure. Such concepts may be adopted from the human system or identified by computational methods. 
Our results showed examples in which the incorporation of early conceptual structure improved the model's learning in terms of accuracy and the amount of data used for training. In addition to these improvements in the learning process, our results, together with previous experiments, suggest that in the learning of a new domain, the early human conceptual structure contributes to what can be considered as understanding of the new domain exhibited by the model.    

An example of such an improved understanding comes from studies related to the Woodward task. Experiments on distinguishing animate and non-animate agents have shown that self-propulsion is a strong early cue for animate agency \citep{premack1990infant}. This property was used in a later study, in which infants at five months of age were shown a box that started to move across a table on its own (self-propulsion) \citep{luo2005can}. Subsequently, they were shown a version of the Woodward test, where the moving agent was the box that they had seen previously. The test showed that the infants treated the box as an animated agent with the goal of reaching its preferred object. These results show that infants can identify novel objects as animate, and they link this attribute with the agent. The results further show that infants follow the implication of the agent’s attribute, leading them to expect the box to show a preference for object-following. 
In contrast, the large models we examined (Table \ref{tab:vlm_results}), did not exhibit a similar association between identifying an entity as an animate agent and following the implication of this identification in terms of predicting the expected agent behavior. We tested the eight large models (in Table \ref{tab:vlm_results}) on their ability to identify the agents we used in our studies as either animate or non-animate, and found that all the models reached high accuracy in this task (average identification accuracy of 98\%). At the same time, for six of the eight models, the accuracy in correctly predicting the behavior of animate agents was low (between 35\%-53\%), indicating that, unlike infants, they did not follow the animacy implication.  

In future research, it would be of interest to study the potential benefits of incorporating aspects of early human conceptual structures into network models. Many of the early conceptual structures acquired by infants are based on visual learning, and this early learning develops before substantial language acquisition. This suggests that attempts to incorporate early concepts into vision-language models could be directed, at least in part, toward the visual components of the models. This suggestion is also supported by previous studies, which have shown that the recognition of sophisticated relationships and interactions occurs in the visual system \citep{hafri2021perception}, and have found similarities between human and model processing of social interactions using videos similar to ours \citep{schiatti2025exploring}. It will therefore be of interest to train the visual part of vision-language models on early visual tasks and basic concepts that are incorporated into the human system. Following training, it will be possible to test for possible improvements in the models. Improvements can fall into two categories: the first is improvement of the models’ capacity in different tasks, and the second is making the models more aligned with human cognition. If successful, the incorporation of early conceptual structures in visual processing could contribute to both the study of human vision and the improvement of future computer vision and multimodal models. 

\section*{Declarations}
\textbf{Funding}
This work was supported by an Ariane de Rothschild
Women Doctoral Funding awarded to Shify Treger. Additional support was provided by research funds from the Weizmann Institute to Shimon Ullman.
\textbf{Competing Interests}  
The authors declare no competing interests.
\textbf{Ethics Approval}  
The human participant study was approved by the Institutional Review Board (IRB) of the Weizmann Institute of Science, Rehovot, Israel. All procedures were carried out in accordance with the relevant guidelines and regulations.
\textbf{Informed Consent}  
Informed consent was obtained from all participants prior to participation in the study.
\textbf{Consent for Publication}  
A preliminary version of parts of this work appeared in the Proceedings of the Annual Meeting of the Cognitive Science Society (2025). 
\textbf{Data Availability}
The datasets will be made publicly available upon acceptance of the article.
\textbf{Materials Availability}  
Not applicable.
\textbf{Code Availability}
The code used for the experiments will be made publicly available upon acceptance of the article.
\textbf{Author Contributions}
S.T. and S.U. conceived and designed the research; S.T. conducted the experiments; S.T. and S.U. analyzed the data and wrote the paper; S.U. supervised the research.

\begin{appendices}

\section{Experiments}\label{app_experiments}

\subsection{Technical Details}\label{app_technical}

All experiments were conducted on up to four NVIDIA RTX-6000 GPUs, each with 24GB of RAM. 

In the Scene Representation stage, we employed the AdamW optimizer \cite{loshchilov2017decoupled} with a learning rate of $2 \times 10^{-5}$ and a weight decay of 0.01, using a batch size of 10. 

For the Prediction Generation stage, we used the Adam optimizer \cite{kingma2014adam} with a learning rate of $10^{-6}$; other optimization parameters were set to the model defaults. Training in this stage was conducted with a batch size of 40. 

Hyperparameters were selected without an exhaustive tuning procedure. We did not explicitly fix a random seed, allowing the computational environment to generate a seed independently for each run.

\subsection{Experiment 4: Goal Attribution}\label{app_goal}

\begin{figure}[H]
\centering
\includegraphics[width=\columnwidth]{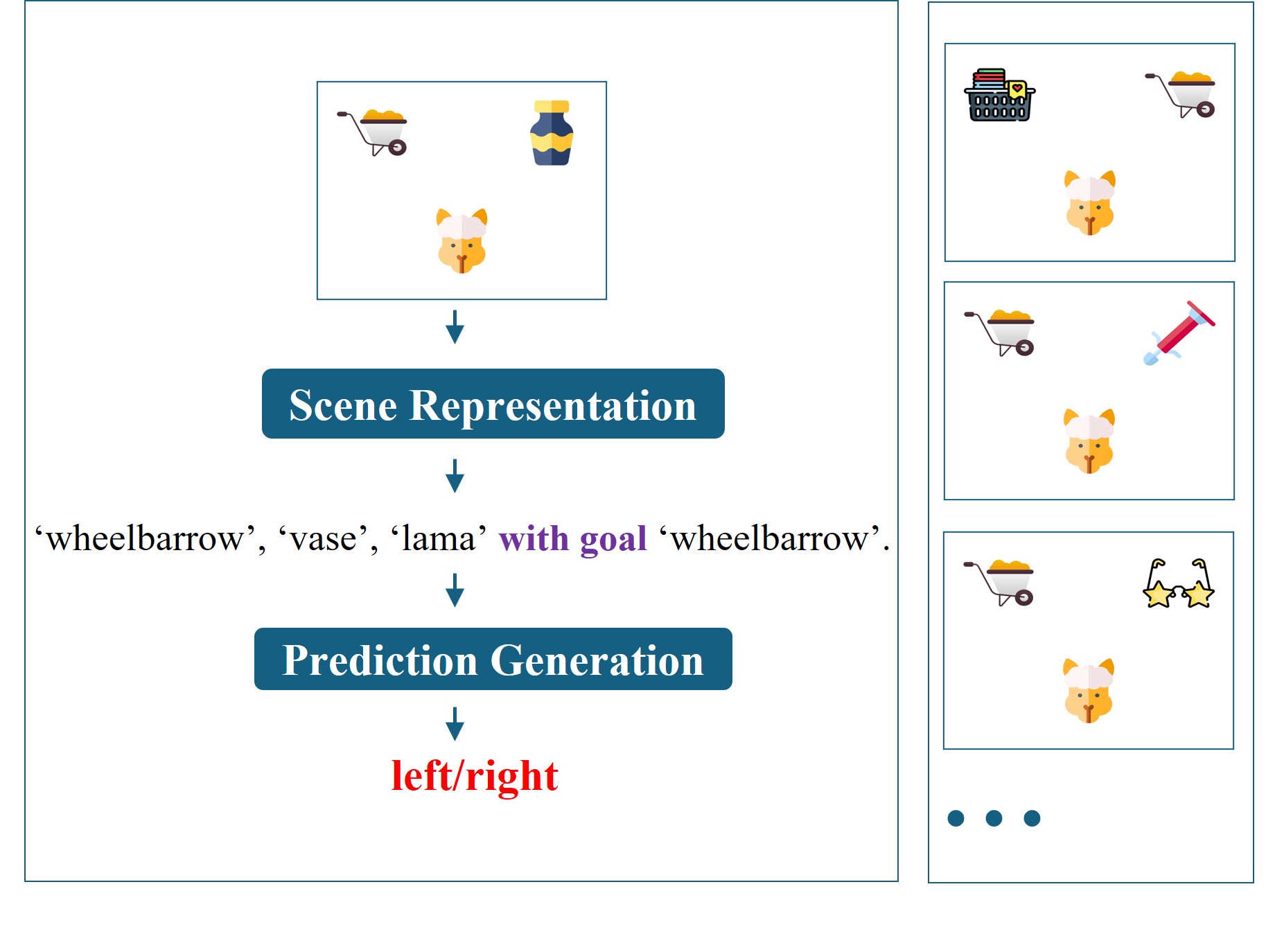}
\caption{Example representation (left) and training data (right) from the goal attribution experiment. The cognitive concept is highlighted in bold and purple. In the training data, each goal-directed actor appears in 10 repetitions with varying distractors across examples.}
\label{goal_app}
\end{figure}

Figure \ref{goal_app} illustrates the representation and training data used in the goal attribution experiment.

\section{Human Experiment}\label{app_human_exp}

\subsection{Participants}

A total of $N = 200$ participants were recruited via Prolific, with 100 participants assigned to each condition (animate and non-animate). 

Each participant completed a single trial only, and no stimulus was repeated within a condition. Participants reported normal or corrected-to-normal vision, fluency in English, and completion of high school (or equivalent education).

\subsection{Procedure}

The experiment was conducted via Prolific. Participants were required to confirm normal or corrected-to-normal vision, completion of high school (or an equivalent educational program), fluency in English, and no prior participation in any of the experiments reported in this paper. 

To ensure data quality, we included two attention checks:  
(1) participants were required to type a specific randomly generated number into a text box;  
(2) following the main task, participants were asked to name at least one object that appeared in the video.  

Only two participants were excluded for failing to complete the task.

\subsection{Stimuli for Human Experiment and VLM Testing}

Stimuli consisted of icon-based visual scenes. Icons were obtained from the Flaticon website under appropriate licensing terms, and additional assets were generated using GPT- and Gemini-based image-generation interfaces.

\section{VLM Testing}\label{app_vlm_details}

\subsection{Statistical Analysis Details}

\begin{table*}[t]

\begin{tabular}{lcccc}
\toprule
& Gemini 3.1 Flash & Gemini 3.1 Pro & Claude 3.5 Sonnet & Claude Opus 4.6 \\
\midrule
Animate & $1.11 \times 10^{-6}$ & 0.0666 & 0.00119 & $6.19 \times 10^{-8}$ \\
Non-Animate & $5.48 \times 10^{-7}$ & 0.0707 & $4.00 \times 10^{-6}$ & 0.00544 \\
\botrule
\end{tabular}

\vspace{6pt}

\begin{tabular}{lcccc}
\toprule
& GPT-4o & GPT-5.2 & Qwen VL Max & Qwen 3.5 Plus \\
\midrule
Animate & $1.31 \times 10^{-8}$ & 0.1149 & $8.73 \times 10^{-5}$ & 0.0771 \\
Non-animate & 0.00113 & $4.12 \times 10^{-4}$ & $2.10 \times 10^{-6}$ & 0.1714 \\
\botrule
\end{tabular}

\caption{P-values from two-proportion z-tests comparing each model’s accuracy to human performance.}
\label{tab:app_vlm_p}

\end{table*}

Human accuracy in each condition was computed over $N = 100$ independent participants (one trial per participant). Model accuracy was computed over the same 100 stimuli per condition.

Because each participant contributed a single independent response and each stimulus was evaluated once per model, observations were treated as independent for the purpose of the proportion tests.

Two-proportion z-tests (pooled variance) were conducted in Python using the \textit{statsmodels} library. All tests were two-tailed with $\alpha = 0.05$.

\subsection{VLM Evaluation Protocol}

Vision–language models were evaluated using a single-call inference protocol. Most models were accessed via the OpenRouter API using an OpenAI-compatible interface, while Gemini models were accessed via their native provider interface.

Temperature was set to 0.0. 

For each trial, video frames were sampled at a rate of 2 frames per second (FPS) across the full duration of the stimulus (up to a maximum of 10 frames per trial) and provided to the model as base64-encoded JPEG images.

Each model input consisted of the following components in fixed order:

1) Introductory task text explaining that the model would view frames from a short video. 2) The sampled video frames. 3) A prediction instruction asking the model to choose which of two candidate images would occur next. 4) The two candidate continuation images (A and B). 5) An explicit reminder of the required output format.

The prediction instruction explicitly required the model to respond using the keys prediction:, explanation:, and memory:, each on a separate line. The model was instructed to select either A or B as the prediction, provide a brief explanation of its reasoning, and name at least one object that appeared in the video.

Model outputs were parsed automatically using rule-based pattern matching to extract the predicted choice (A or B), the explanation text, and the memory response. The maximum generation length was set to 8192 tokens.

\end{appendices}



\begin{thebibliography}{45}
\providecommand{\natexlab}[1]{#1}
\providecommand{\url}[1]{{#1}}
\providecommand{\urlprefix}{URL }
\providecommand{\doi}[1]{\url{https://doi.org/#1}}
\providecommand{\eprint}[2][]{\url{#2}}
 \bibcommenthead

\bibitem[{Baillargeon(1987)}]{baillargeon1987object}
Baillargeon R (1987) Object permanence in 3$1/2$-and 4$1/2$-month-old infants. Developmental psychology 23(5):655

\bibitem[{Biro et~al.(2011)Biro, Verschoor, and Coenen}]{biro2011evidence}
Biro S, Verschoor S, Coenen L (2011) Evidence for a unitary goal concept in 12-month-old infants. Developmental science 14(6):1255--1260

\bibitem[{Bortoletto et~al.(2024)Bortoletto, Shi, and Bulling}]{bortoletto2024neural}
Bortoletto M, Shi L, Bulling A (2024) Neural reasoning about agents’ goals, preferences, and actions. In: Proceedings of the AAAI Conference on Artificial Intelligence, pp 456--464

\bibitem[{Chollet(2019)}]{chollet2019measure}
Chollet F (2019) On the measure of intelligence. arXiv preprint arXiv:191101547

\bibitem[{D'Entremont et~al.(1997)D'Entremont, Hains, and Muir}]{d1997demonstration}
D'Entremont B, Hains SM, Muir DW (1997) A demonstration of gaze following in 3-to 6-month-olds. Infant Behavior and Development 20(4):569--572

\bibitem[{Devlin(2018)}]{devlin2018bert}
Devlin J (2018) Bert: Pre-training of deep bidirectional transformers for language understanding. arXiv preprint arXiv:181004805

\bibitem[{Dziri et~al.(2024)Dziri, Lu, Sclar, Li, Jiang, Lin, Welleck, West, Bhagavatula, Le~Bras et~al.}]{dziri2024faith}
Dziri N, Lu X, Sclar M, et~al (2024) Faith and fate: Limits of transformers on compositionality. Advances in Neural Information Processing Systems 36

\bibitem[{Falck-Ytter et~al.(2006)Falck-Ytter, Gredeb{\"a}ck, and Von~Hofsten}]{falck2006infants}
Falck-Ytter T, Gredeb{\"a}ck G, Von~Hofsten C (2006) Infants predict other people's action goals. Nature neuroscience 9(7):878--879

\bibitem[{Flom et~al.(2017)Flom, Lee, and Muir}]{flom2017gaze}
Flom R, Lee K, Muir D (2017) Gaze-following: Its development and significance. Psychology Press

\bibitem[{Friedman and Neary(2008)}]{friedman2008determining}
Friedman O, Neary KR (2008) Determining who owns what: Do children infer ownership from first possession? Cognition 107(3):829--849

\bibitem[{Gandhi et~al.(2021)Gandhi, Stojnic, Lake, and Dillon}]{gandhi2021baby}
Gandhi K, Stojnic G, Lake BM, et~al (2021) Baby intuitions benchmark (bib): Discerning the goals, preferences, and actions of others. Advances in neural information processing systems 34:9963--9976

\bibitem[{Gergely et~al.(1995)Gergely, N{\'a}dasdy, Csibra, and B{\'\i}r{\'o}}]{gergely1995taking}
Gergely G, N{\'a}dasdy Z, Csibra G, et~al (1995) Taking the intentional stance at 12 months of age. Cognition 56(2):165--193

\bibitem[{Gergely et~al.(2002)Gergely, Bekkering, and Kir{\'a}ly}]{gergely2002rational}
Gergely G, Bekkering H, Kir{\'a}ly I (2002) Rational imitation in preverbal infants. Nature 415(6873):755--755

\bibitem[{Hafri and Firestone(2021)}]{hafri2021perception}
Hafri A, Firestone C (2021) The perception of relations. Trends in Cognitive Sciences 25(6):475--492

\bibitem[{Hein and Diepold(2023)}]{hein2023comparing}
Hein A, Diepold K (2023) Comparing intuitions about agents’ goals, preferences and actions in human infants and video transformers. In: Proceedings of the Annual Meeting of the Cognitive Science Society

\bibitem[{Hespos and Baillargeon(2001)}]{hespos2001reasoning}
Hespos SJ, Baillargeon R (2001) Reasoning about containment events in very young infants. Cognition 78(3):207--245

\bibitem[{Kingma(2014)}]{kingma2014adam}
Kingma DP (2014) Adam: A method for stochastic optimization. arXiv preprint arXiv:14126980

\bibitem[{Kir{\'a}ly et~al.(2003)Kir{\'a}ly, Jovanovic, Prinz, Aschersleben, and Gergely}]{kiraly2003early}
Kir{\'a}ly I, Jovanovic B, Prinz W, et~al (2003) The early origins of goal attribution in infancy. Consciousness and cognition 12(4):752--769

\bibitem[{Kirillov et~al.(2023)Kirillov, Mintun, Ravi, Mao, Rolland, Gustafson, Xiao, Whitehead, Berg, Lo et~al.}]{kirillov2023segment}
Kirillov A, Mintun E, Ravi N, et~al (2023) Segment anything. In: Proceedings of the IEEE/CVF International Conference on Computer Vision, pp 4015--4026

\bibitem[{Lake and Baroni(2018)}]{lake2018generalization}
Lake B, Baroni M (2018) Generalization without systematicity: On the compositional skills of sequence-to-sequence recurrent networks. In: International conference on machine learning, PMLR, pp 2873--2882

\bibitem[{Lake et~al.(2017)Lake, Ullman, Tenenbaum, and Gershman}]{lake2017building}
Lake BM, Ullman TD, Tenenbaum JB, et~al (2017) Building machines that learn and think like people. Behavioral and brain sciences 40:e253

\bibitem[{Li et~al.(2022)Li, Li, Xiong, and Hoi}]{li2022blip}
Li J, Li D, Xiong C, et~al (2022) Blip: Bootstrapping language-image pre-training for unified vision-language understanding and generation. In: International conference on machine learning, PMLR, pp 12888--12900

\bibitem[{Li et~al.(2024)Li, Yasuda, Dillon, and Lake}]{li2024infant}
Li W, Yasuda SC, Dillon MR, et~al (2024) An infant-cognition inspired machine benchmark for identifying agency, affiliation, belief, and intention. In: Proceedings of the Annual Meeting of the Cognitive Science Society

\bibitem[{Libertus et~al.(2016)Libertus, Greif, Needham, and Pelphrey}]{libertus2016infants}
Libertus K, Greif ML, Needham AW, et~al (2016) Infants’ observation of tool-use events over the first year of life. Journal of experimental child psychology 152:123--135

\bibitem[{Lin et~al.(2022)Lin, Stavans, and Baillargeon}]{lin2022infants}
Lin Y, Stavans M, Baillargeon R (2022) Infants’ physical reasoning and the cognitive architecture that supports it. Cambridge handbook of cognitive development pp 168--194

\bibitem[{Loshchilov and Hutter(2017)}]{loshchilov2017decoupled}
Loshchilov I, Hutter F (2017) Decoupled weight decay regularization. arXiv preprint arXiv:171105101

\bibitem[{Luo and Baillargeon(2005)}]{luo2005can}
Luo Y, Baillargeon R (2005) Can a self-propelled box have a goal? psychological reasoning in 5-month-old infants. Psychological science 16(8):601--608

\bibitem[{Oh et~al.(2024)Oh, Kim, Cha, and Oh}]{oh2024generative}
Oh J, Kim E, Cha I, et~al (2024) The generative ai paradox on evaluation: What it can solve, it may not evaluate. arXiv preprint arXiv:240206204

\bibitem[{Phillips and Wellman(2005)}]{phillips2005infants}
Phillips AT, Wellman HM (2005) Infants' understanding of object-directed action. Cognition 98(2):137--155

\bibitem[{Phillips et~al.(2002)Phillips, Wellman, and Spelke}]{phillips2002infants}
Phillips AT, Wellman HM, Spelke ES (2002) Infants' ability to connect gaze and emotional expression to intentional action. Cognition 85(1):53--78

\bibitem[{Premack(1990)}]{premack1990infant}
Premack D (1990) The infant's theory of self-propelled objects. Cognition 36(1):1--16

\bibitem[{Saxe et~al.(2005)Saxe, Tenenbaum, and Carey}]{saxe2005secret}
Saxe R, Tenenbaum J, Carey S (2005) Secret agents: Inferences about hidden causes by 10-and 12-month-old infants. Psychological science 16(12):995--1001

\bibitem[{Saylor et~al.(2011)Saylor, Ganea, and V{\'a}zquez}]{saylor2011s}
Saylor MM, Ganea PA, V{\'a}zquez MD (2011) What’s mine is mine: Twelve-month-olds use possessive pronouns to identify referents. Developmental Science 14(4):859--864

\bibitem[{Scaife and Bruner(1975)}]{scaife1975capacity}
Scaife M, Bruner JS (1975) The capacity for joint visual attention in the infant. Nature 253(5489):265--266

\bibitem[{Schiatti et~al.(2025)Schiatti, Vallarino, Lopez, Kuo, Moro, Zhang, Gori, Del~Bue, Katz, and Barbu}]{schiatti2025exploring}
Schiatti L, Vallarino G, Lopez SM, et~al (2025) Exploring human-model alignment in visual social attention during help-and-hinder social interaction classification. In: Proceedings of the IEEE/CVF International Conference on Computer Vision, pp 4854--4863

\bibitem[{Spelke(2022)}]{spelke2022babies}
Spelke ES (2022) What babies know: Core Knowledge and Composition volume 1, vol~1. Oxford University Press

\bibitem[{Spelke et~al.(1995)Spelke, Kestenbaum, Simons, and Wein}]{spelke1995spatiotemporal}
Spelke ES, Kestenbaum R, Simons DJ, et~al (1995) Spatiotemporal continuity, smoothness of motion and object identity in infancy. British journal of developmental psychology 13(2):113--142

\bibitem[{Stojni{\'c} et~al.(2023)Stojni{\'c}, Gandhi, Yasuda, Lake, and Dillon}]{stojnic2023commonsense}
Stojni{\'c} G, Gandhi K, Yasuda S, et~al (2023) Commonsense psychology in human infants and machines. Cognition 235:105406

\bibitem[{Tomasello(2009)}]{tomasello2009cultural}
Tomasello M (2009) The cultural origins of human cognition. Harvard university press

\bibitem[{Ullman et~al.(2019)Ullman, Dorfman, and Harari}]{ullman2019model}
Ullman S, Dorfman N, Harari D (2019) A model for discovering ‘containment’relations. Cognition 183:67--81

\bibitem[{Valle-Perez et~al.(2018)Valle-Perez, Camargo, and Louis}]{valle2018deep}
Valle-Perez G, Camargo CQ, Louis AA (2018) Deep learning generalizes because the parameter-function map is biased towards simple functions. arXiv preprint arXiv:180508522

\bibitem[{Woodward(1998)}]{woodward1998infants}
Woodward AL (1998) Infants selectively encode the goal object of an actor's reach. Cognition 69(1):1--34

\bibitem[{Woodward(1999)}]{woodward1999infants}
Woodward AL (1999) Infants’ ability to distinguish between purposeful and non-purposeful behaviors. Infant behavior and development 22(2):145--160

\bibitem[{Woodward and Sommerville(2000)}]{woodward2000twelve}
Woodward AL, Sommerville JA (2000) Twelve-month-old infants interpret action in context. Psychological Science 11(1):73--77

\bibitem[{Zhi-Xuan et~al.(2022)Zhi-Xuan, Gothoskar, Pollok, Gutfreund, Tenenbaum, and Mansinghka}]{zhi2022solving}
Zhi-Xuan T, Gothoskar N, Pollok F, et~al (2022) Solving the baby intuitions benchmark with a hierarchically bayesian theory of mind. arXiv preprint arXiv:220802914

\end{thebibliography}
\end{document}